# Matryoshka Networks: Predicting 3D Geometry via Nested Shape Layers

Stephan R. Richter[1,2*]   Stefan Roth[1]

[1]TU Darmstadt   [2] Intel Labs

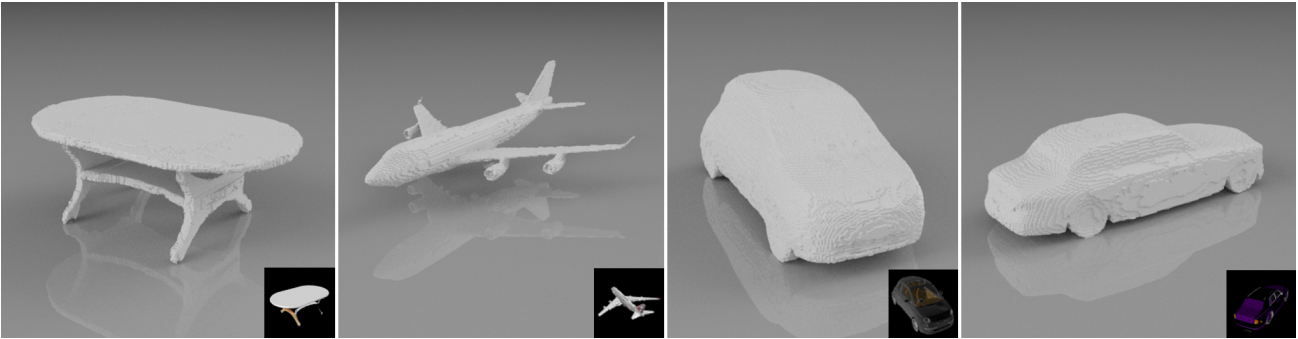

Figure 1. **Exemplary shape reconstructions** from a single image by our *Matryoshka network* based on nested shape layers.

## Abstract

*In this paper, we develop novel, efficient 2D encodings for 3D geometry, which enable reconstructing full 3D shapes from a single image at high resolution. The key idea is to pose 3D shape reconstruction as a 2D prediction problem. To that end, we first develop a simple baseline network that predicts entire* voxel tubes *at each pixel of a reference view. By leveraging well-proven architectures for 2D pixel-prediction tasks, we attain state-of-the-art results, clearly outperforming purely voxel-based approaches. We scale this baseline to higher resolutions by proposing a memory-efficient shape encoding, which recursively decomposes a 3D shape into nested shape layers, similar to the pieces of a* Matryoshka doll*. This allows reconstructing highly detailed shapes with complex topology, as demonstrated in extensive experiments; we clearly outperform previous octree-based approaches despite having a much simpler architecture using standard network components. Our Matryoshka networks further enable reconstructing shapes from IDs or shape similarity, as well as shape sampling.*

## 1. Introduction

Being able to reason about the 3D shape of objects, even when presented with only a single monocular image, is one of the remarkable abilities of the human visual system. In the absence of geometric cues from stereopsis or motion, our visual system is still able to infer detailed surfaces or plausibly complete hidden parts.

The advent of large-scale shape collections [4] and advances in data-driven approaches, especially convolutional neural networks (CNNs), have sparked new interest in developing approaches that mimic the human visual system in its ability to reconstruct 3D shapes from a single image, *e.g.* [5, 15, 20, 22, 27, 31]. The predominant structure of CNNs employed for this task is an hourglass shape with an encoder, which transforms a single image into a shape code, and a decoder finally producing a 3D shape [5, 10]. Interpreting the shape code as a multi-dimensional tensor with spatial and feature dimensions, the decoder successively increases the spatial resolution of the shape code while reducing the number of feature dimensions. The output of the decoder is a volumetric binary occupancy map. The overall down-sampling and up-sampling of representations in this hourglass architecture facilitates the accumulation of shape information from the whole image and propagating it to all parts of the reconstructed shape. Higher resolutions of the input and/or output require more levels of scaling, which results in deeper networks. The network depth is in turn bound by the available GPU memory, impeding CNNs with volumetric decoders in their ability to reconstruct shapes at high resolution [11, 22, 27].

More efficient encodings, for example based on octrees [11, 22, 27], alleviate this problem, but require sophisticated structures and mechanisms for feature propagation through the decoder, impeding portability across deep leaning frameworks and exploration of alternative architectures. Alternatively, view-based reconstructions [18, 25] can encode highly detailed shape information, but cannot represent shapes with a significant level of self-occlusion.

---
*This work was carried out while at TU Darmstadt.

In this paper, we develop a novel, efficient 2D encoding for 3D geometry, which enables reconstructing full 3D shapes from a single image at high resolution. We begin by developing a new architecture for dense 3D shape reconstruction at low resolutions. Its key feature is that we pose reconstructing 3D voxel occupancy as a 2D prediction problem by directly predicting whole *voxel tubes* at every pixel of a reference view. This allows us to use a wide range of standard networks for 2D pixel-prediction tasks, which enables this simple baseline to attain state-of-the-art accuracy, clearly outperforming previous purely voxel-based approaches. Another factor in reaching such high accuracy levels is using a structured loss function.

We then scale this baseline to higher resolutions by proposing an efficient shape encoding based on the idea of *nested shape layers*. That is, the object shape is recursively decomposed into nested layers, similar to the pieces of a *Matryoshka doll*, see Fig. 4. This has several key advantages: *(1)* it allows for a highly detailed reconstruction of shapes with complex topology, including self-occlusions; *(2)* each shape layer can be represented through a set of six depth maps, which is memory efficient and allows the use of standard network architectures; *(3)* nested shape layers lead to more detailed reconstructions than octree-based architectures despite being much simpler. We further demonstrate the capabilities of the proposed encoding and decoder architecture in reconstructing shapes from IDs or shape similarity, as well as in shape sampling.

## 2. Prior Work

With the advent of large-scale shape collections [4], data-driven methods, and especially CNNs, have become the method of choice for predicting 3D shapes. Inspired by the success of CNNs for dense 2D prediction tasks, Wu *et al.* [31] adapted CNNs to volumetric outputs. Yan *et al.* [33] and Zhu *et al.* [34] showed that optimizing projections of the predicted shape benefits the reconstruction. Choy *et al.* [5] developed a joint approach for shape reconstruction from one or multiple views. Girdhar *et al.* [10] combined an autoencoder and a convolutional network to learn an embedding of images and 3D shapes. Wu *et al.* [30] trained a generative adversarial network to synthesize 3D shapes. Tulsiani *et al.* [28] learned a shape decoder from 2D supervision. Wu *et al.* [29] used intermediate 2.5D shape representations in order to decouple image encoding and 3D shape decoding. All have in common that they model 3D shapes as binary occupancy maps. This allows for casting shape estimation as a classification problem at the voxel level and benefitting from the extraordinary performance of CNNs in classification tasks. Representing each voxel separately to facilitate the classification task comes at a price, however, as the memory requirements scale cubically with the resolution of shapes. Consequently, the output resolution is usually limited to 32 voxels along each side.

Riegler *et al.* [22] addressed the memory requirements of predicting high-resolution occupancy maps by adapting CNNs to operate on octrees. However, their method requires the tree structure to be known ahead of time, which limits its applicability for 3D reconstruction. The works of Tatarchenko *et al.* [27] and Häne *et al.* [11] alleviate this problem by also predicting the tree structure. Besides commonly requiring custom network layers [22, 27], which impedes porting the approaches to other deep learning libraries, the sparse structure of octrees complicates feature propagation to neighboring cells. This is in contrast to the proposed method, which only requires network layers that are standard in all common deep learning frameworks and, by building on 2D convolutions, facilitates the easy exploration of recent advances in network architectures.

Fan *et al.* [8] addressed the sparse structure of shapes within a 3D volume by explicitly predicting a point cloud. Their method demonstrated impressive results at low resolution, but it has yet to be seen if and how well this approach scales to higher resolutions.

As an alternative to a volumetric representation, Tatarchenko *et al.* [26] trained a CNN to generate RGB-D images from arbitrary views of an object. In a post-processing step, the different views are merged into a consistent shape. Following this approach for the generation of shapes, Soltani *et al.* [25] predicted pairs of silhouettes and depth images for a fixed set of views, and Lun *et al.* [18] additionally predicted surface normal maps. The final fusion of views has been addressed by merging them into a point cloud and pruning outliers using the predicted silhouettes [25], registering views and solving an optimization problem [18], as well as learning a differentiable depth map renderer to produce consistent projections [17]. In general, view-based methods are able to generate shapes at high resolutions, but occasionally suffer from noisy estimates, which need to be addressed in the fusion step. Furthermore, view-based methods cannot handle large self-occlusions. Our proposed method addresses the fusion step and handling of occlusions in a simple, but efficient formulation.

Sinha *et al.* [24] projected object surfaces to geometry images in order to build on image-based CNN architectures. This shape representation allows for a very memory-efficient encoding, but requires additional care for handling different mesh topologies and projective distortions produce a non-uniformly distributed level of detail. Zou *et al.* [35] assembled 3D shapes from volumetric primitives. Our work, in contrast, is inspired by depth peeling [7] and constructive solid geometry [13]. Gallup *et al.* [9] used a $n$-layer heightmap representation to constrain the reconstructions of buildings from occupancy grids. Our nested shape layers can be seen as a generalization of this representation as our Matryoshka networks effectively estimate 6 overlapping heightmaps per layer, which are fused together.

In concurrent work, Delanoy *et al.* [6] explored the prediction of multi-channel depth maps in the context of reconstructing aligned shapes from sketches, which allows for exploiting additional projective constraints.

## 3. Formulation

We develop a framework for memory-efficient prediction of 3D shapes in two stages. First, we encode 3D shapes as $n$-channel images, where each pixel represents a tube of $n$ voxels in a 3-dimensional tensor (a fiber along the $z$-axis). This leads to a more memory-efficient intermediate representation, since features are shared across entire fibers instead of a single voxel (cell). To that end, we adapt network architectures for dense pixel-prediction tasks to predicting voxel tubes. This reduces the memory footprint of the network, but still produces a dense binary occupancy map in the last network layer. Second, we further compress the output by predicting nested shape layers that can encode shapes with arbitrary amounts of self-occlusion. Each shape layer consists of 6 depth maps, *c.f.* Fig. 4. Through careful alignment of the depth maps and an appropriate loss function, we avoid noisy estimates, a costly fusion via optimization, and minimize the dimensionality of the final network layer.

### 3.1. Standard voxel decoder

To ground our discussions, let us begin by describing a simple standard architecture for predicting volumetric binary occupancy maps, as it is common to a wide range of previous work [5, 10, 33]. We here focus our discussion on the decoder and assume that its input, a shape code $\mathbb{S}$ with a spatial resolution of $n_s \times n_s \times n_s$ and $n_f$ features, is provided by an image encoder. Each layer of the voxel decoder then up-samples the shape code, *i.e.* it increases the spatial resolution while decreasing the number of features until the full spatial resolution $n_o \times n_o \times n_o$ has been reached and only one feature dimension is left. The resulting 4-dimensional tensor is then interpreted as a 3D binary occupancy map. Intuitively, the voxel decoder transforms a 4-dimensional tensor into 3-dimensional tensor by iteratively lowering the feature resolution. Fig. 2 (left) shows an illustration.

### 3.2. Predicting voxel tubes

To address the memory inefficiency of such a standard voxel decoder, we here propose to predict entire voxel tubes. The key idea is that we interpret the shape code as a 3-dimensional tensor with spatial dimensions $n_t \times n_t$ and one feature dimension. Analogously to the voxel decoder, we up-sample the spatial resolution while down-sampling the number of features. Different to the voxel decoder, however, we reduce the number of features until it equals $n_o$. Hence, the output of our decoder is a 3-dimensional tensor with resolution of $n_o \times n_o \times n_o$. With this simple change in representation, a fiber of features no longer encodes a single voxel, but a complete tube of voxels jointly. Therefore, we term the resulting tensor a *voxel tube image*. Fig. 2 (middle) illustrates the architecture. As the proposed decoder now operates on images instead of voxel grids, we can employ standard 2D network components for designing the decoder and take full advantage of recent advances in architectures for 2D prediction tasks [12, 14, 23, 32].

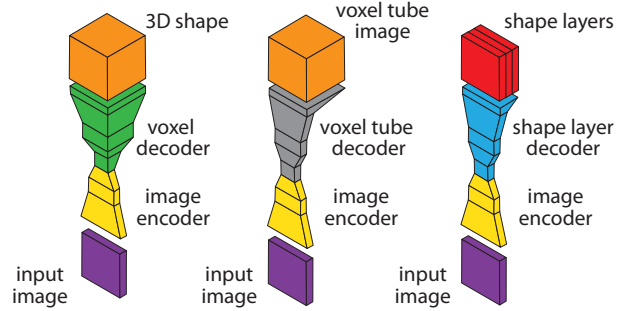

Figure 2. **Memory-efficient geometry decoders.** Encoding features jointly per voxel tube turns a standard voxel decoder (left) into a *voxel tube image* decoder (middle). Predicting *shape layers* (right) allows for reconstructing shapes at higher resolution.

### 3.3. Shape layers: Learning to compress voxel tubes

Although sharing features across voxel tubes reduces the space requirements of the decoder, it is insufficient for scaling the output resolution by multiple octaves with currently available GPU architectures. To scale our approach to higher resolutions, we compress shapes by constructing them from multiple *shape layers*, each of which requires only $n_o \times n_o \times 6$ activations in the network output. Each shape layer $S \in \mathcal{S}$ is the product of fusing 6 depth maps $\mathbf{d} = (d_{-x}, d_{+x}, d_{-y}, d_{+y}, d_{-z}, d_{+z}) \in \mathcal{D}$, which represent a shape as depicted in Fig. 3 (left).

Specifically, each depth map $d_i$ is an orthogonal projection imaged from view position $v_i$, which is located at the center of side $i$ of an axis-aligned unit cube. We assume the cube to be at the origin, and all views to face the origin.

**Shape from depth maps.** For each of the three axes, we now define shapes as

$$S_x \equiv \{(i,j,k) \mid d_{-x}(j,k) \leq i \leq n_o - d_{+x}(j,k)\} \quad \text{(1a)}$$
$$S_y \equiv \{(i,j,k) \mid d_{-y}(i,k) \leq j \leq n_o - d_{+y}(i,k)\} \quad \text{(1b)}$$
$$S_z \equiv \{(i,j,k) \mid d_{-z}(i,j) \leq k \leq n_o - d_{+z}(i,j)\}, \quad \text{(1c)}$$

where the tuple $(i,j,k)$ indexes a cell in a binary occupancy map of size $n_o \times n_o \times n_o$. That is, the shape $S_x$, for example, can be thought of as being represented by sending $x$-axis-aligned rays through it and recording where they enter the shape and exit again. Put differently, the pairs $(d_{-x}, d_{+x})$, $(d_{-y}, d_{+y})$, and $(d_{-z}, d_{+z})$ are effectively run-length encodings of the shapes $S_x$, $S_y$, and $S_z$. The colored car shapes in Fig. 3 (right) illustrate this. Note how a single shape, say $S_z$, is not sufficient to represent the car (shown in yellow), since the geometry of the wheels cannot be recovered correctly from this view due to self-occlusion.

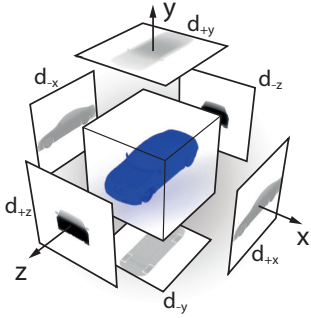 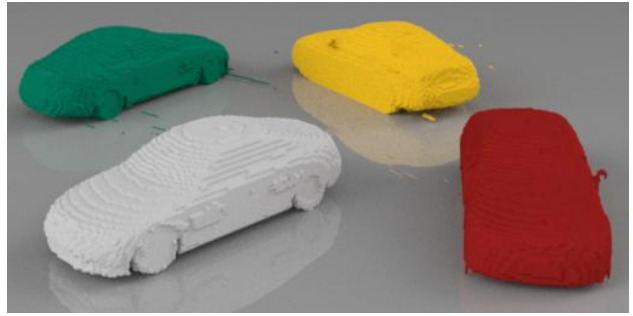

Figure 3. **Fusion of depth maps**. We interpret pairs of depth maps taken along three view axes (left) as run-length encoding of geometry and fuse them to shapes $S_x$ (right, green), $S_y$ (red), and $S_z$ (yellow). Noisy predictions cause a smearing at the shape silhouettes. By intersecting $S_x$, $S_y$, and $S_z$, we obtain a shape $S$ (white) with outliers removed.

**Shape fusion.** We address this by fusing the three shapes via their intersection as

$$S = \phi(\mathbf{d}) \equiv S_x \cap S_y \cap S_z \quad \text{with} \quad \phi : \mathcal{D} \to \mathcal{S}. \quad (2)$$

The result is shown as the white car shape in Fig. 3 (right). This fusion process and the placement of the three orthogonal views $v_i$ is motivated by our observation that depth map predictions are often less accurate near the silhouette of an object. This is intuitive as the decision whether to assign a pixel to foreground or background is less certain close to the silhouette. If we cast the occupancy prediction as a classification problem (*c.f.* Sec. 3.2), we can assess the uncertainty through the softmax predictions. Depth map prediction is cast as regression problem here, however, and the network tends to average multiple plausible predictions. This has been observed also for point clouds [8]. In regions around the silhouette, this averaging causes noisy estimates and a smearing of the shape as can be seen in the colored shapes in Fig. 3 (right). By placing orthogonal views $v_i$ at the sides of a unit cube, we ensure that regions of high uncertainty in one view are complemented by regions of low uncertainty in another. Fusing the shapes $S_x$, $S_y$, and $S_z$ through their intersection thus allows to remove outliers reliably.

### 3.4. Nested shape layers: Recovering occluded parts

Representing shapes through a single set of 6 depth maps cannot possibly recover parts that are occluded from all three view axes. We address this by building up a shape $S_{1:L}$ from $L$ *nested shape layers* by iteratively adding and subtracting shapes $\phi(\mathbf{d}_i)$. This process is inspired by constructive solid geometry [13]. Let $\phi : \mathcal{D} \to \mathcal{S}$ be the fusion of a shape from the set of depth maps as defined in Eq. (2). We then compose shapes via the recursion

$$S_1 \equiv \phi(\mathbf{d}_1) \quad (3a)$$
$$S_{1:2n} \equiv S_{1:2n-1} \setminus \phi(\mathbf{d}_{2n}) \quad (3b)$$
$$S_{1:2n+1} \equiv S_{1:2n} \cup \phi(\mathbf{d}_{2n+1}), \quad (3c)$$

where $n \in \mathbb{N}^+$. We begin the recursion by only fusing depth maps in the first layer (Eq. 3a), then we subtract shapes in even layers (Eq. 3b), and add shapes in odd layers (Eq. 3c). This process allows us to encode complex geometries; Fig. 4 shows an exemplary encoding of a shape into multiple nested shape layers. Note that the nesting of the shape layers is akin to *Matryoshka dolls* (Fig. 4, right), *i.e.* the first two shape layers encode the outermost doll, the next two layers describe the second doll inside the first, and so on.

**Learning.** To learn to predict nested shape layers, we need to define the appropriate ground truth depth maps. To that end, let $T_{1:L} \in \mathcal{S}$ be the (true) target shape and $\pi : \mathcal{S} \to \phi(\mathcal{D})$ be the projection from an arbitrary shape to the space of shapes that can be represented by the depth map fusion process $\phi$ from Eq. (2). To compute the projection, we greedily apply a simple ray casting and store the depth of the first intersection with the shape. The ground truth is then given by the recursion

$$T_1 \equiv \pi(T_{1:L}) \quad (4a)$$
$$T_{1:2n} \equiv \pi(T_{1:2n-1} \setminus T_{1:L}) \quad (4b)$$
$$T_{1:2n+1} \equiv \pi(T_{1:L} \setminus T_{1:2n}). \quad (4c)$$

Note that although the shapes are encoded recursively, we train a single network to predict all layers jointly.

**How many shape layers do we need?** To answer this question, we encode shapes from the ShapeNet-core dataset [4] with a varying number $L$ of layers and compute the intersection over union between shapes before encoding ($T_{1:L}$)

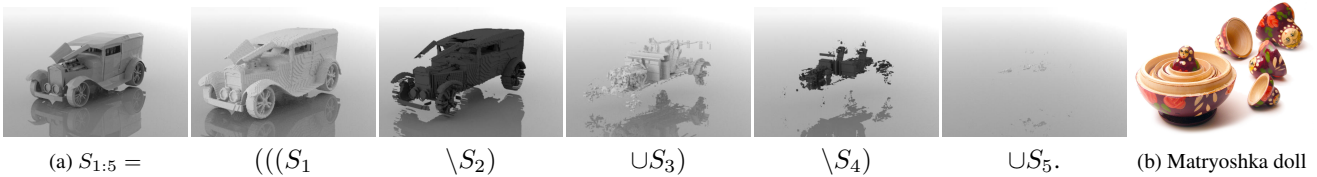

(a) $S_{1:5} =$  $(((S_1$  $\setminus S_2)$  $\cup S_3)$  $\setminus S_4)$  $\cup S_5.$  (b) Matryoshka doll

Figure 4. **Composing shapes from nested shape layers.** The proposed method reconstructs a shape $S_{1:5}$ by iteratively adding ($S_1$,$S_3$,$S_5$) and subtracting ($S_2$,$S_4$) shape layers built from fused depth maps (a). This is akin to the layers of a Matryoshka doll (b).

and after decoding ($S_{1:L}$). We report results in Tab. 1. We find that 94.8% of shapes at a low resolution ($32^3$, provided by Choy et al. [5]) can be completely encoded with a single shape layer and only 4 shapes require more than 2 layers. Evaluating shapes from ShapeNet-cars at $128^3$ (from Tatarchenko et al. [27]), we find that only 2.6% of shapes can be completely encoded with just a single shape layer. This demonstrates the need for a nested representation to accurately represent shapes at high resolution.

### 3.5. Loss functions for dense and sparse prediction

As a voxel can either be occupied or empty, the prediction of occupancies within a voxel grid is often cast as binary classification, minimizing the binary cross entropy [5, 10, 22, 27]. This is in contrast to the metrics commonly used for evaluating predictions [2]. Most common is the intersection over union (IoU), or Jaccard index

$$\text{IoU}(A, B) = \frac{|A \cap B|}{|A \cup B|}. \quad (5)$$

The IoU divides number of true positives (the intersection) by the sum of true positives, false positives, and false negatives (the union). In the context of segmenting a 3D object, correct foreground predictions are thus effectively weighted by the size of the true object and the prediction. Consequently, the contribution of a single voxel toward the overall loss depends on the remaining predictions within the voxel grid. This is in contrast to the binary cross entropy or typical regression losses, e.g. $\mathcal{L}_1$ or $\mathcal{L}_2$, which decompose into losses of individual voxels (or pixels).

| Number of shape layers | 1 | 2 | 3 | 4 | 5 |
|---|---|---|---|---|---|
| *ShapeNet-core* $32^3$ | | | | | |
| Completely reconstructed | 94.8 | 100.0 | 100.0 | 100.0 | 100.0 |
| Mean IoU of reconstruction | 99.9 | 100.0 | 100.0 | 100.0 | 100.0 |
| *ShapeNet-cars* $128^3$ | | | | | |
| Completely reconstructed | 2.6 | 35.2 | 94.3 | 99.9 | 100.0 |
| Mean IoU of reconstruction | 97.8 | 99.9 | 100.0 | 100.0 | 100.0 |

Table 1. **Modeling power of nested shape layers.** Percentage of ShapeNet-core/cars shapes completely reconstructed with given number of shape layers. Higher resolutions require more layers.

Alternatively, we also consider the cosine similarity

$$C(A, B) = \frac{A \cdot B}{\|A\|_2 \|B\|_2}, \quad (6)$$

which has been used for learning embeddings, e.g. [3], but as far as we know not in a reconstruction setting. To adapt cosine similarity and IoU to our setting, c.f. [1, 2], we define

$$\mathcal{L}_C(\bar{\mathbf{x}}, \bar{\mathbf{y}}) = 1 - \langle \bar{\mathbf{x}}, \bar{\mathbf{y}} \rangle \quad (7)$$

$$\mathcal{L}_{\text{IoU}}(\bar{\mathbf{x}}, \bar{\mathbf{y}}) = 1 - \frac{\langle \bar{\mathbf{x}}, \bar{\mathbf{y}} \rangle}{\sum_i \bar{\mathbf{x}}_i + \bar{\mathbf{y}}_i - \bar{\mathbf{x}}_i \bar{\mathbf{y}}_i}, \quad (8)$$

where $\bar{\mathbf{x}}$, $\bar{\mathbf{y}}$ are predicted and ground truth shape each stacked into a vector and normalized to unit norm. Note that $\bar{\mathbf{x}}$ is based directly on the softmax outputs.

**Loss functions for shape layers.** Employing $\mathcal{L}_C$ and $\mathcal{L}_{\text{IoU}}$ (Eqs. 7 and 8) for predicting (nested) shape layers would require decoding the representation into a voxel grid during training, thus counteracting the efficiency gains from the compact representation. We hence opt for a different loss function for training the prediction of (nested) shape layers. Estimating depth (or run-lengths) is naturally a regression task, which we address via a robust $\mathcal{L}_1$-penalty. Applying a regression loss naïvely to the full depth map, however, will bias the network toward background pixels. This has been addressed in the literature by predicting separate foreground masks [18, 25], requiring an additional channel per depth map. Avoiding auxiliary outputs, we modify the employed loss to adaptively weigh foreground and background regions by computing the average loss separately for foreground and background regions. We further refrain from forcing background pixels to equal any specific value, as this would unnecessarily bind model capacity. Thus, we require background pixels to take on values less than a margin $m$, instead. Our modified loss for each pixel then becomes

$$\mathcal{L}'_1(x, y) = \begin{cases} |x - y|, & \text{if } y > 0 \\ \max(0, x + m), & \text{otherwise}, \end{cases} \quad (9)$$

where $x$ and $y$ are prediction and label for a pixel. We also experimented with the $\mathcal{L}_2$-norm as basis for our modification, but observed significantly worse reconstructions.

### 3.6. Implementation details

Our networks can be structured into an encoder and a decoder with a bottleneck in the middle (Fig. 2, middle &

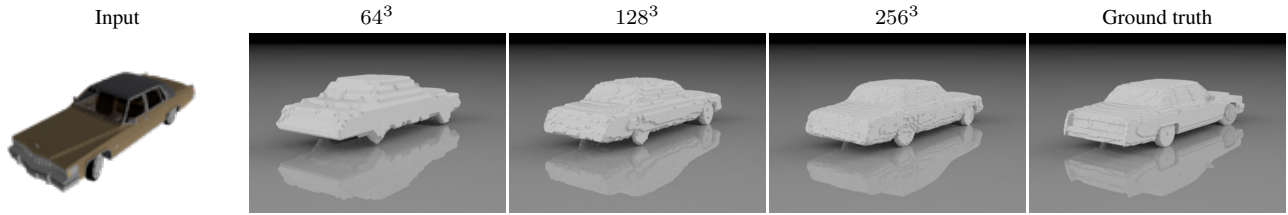

Figure 5. **Shapes reconstructed from a single image by our Matryoshka network at different resolutions.**

| Method | airplane | bench | cabinet | car | cellphone | chair | couch | firearm | lamp | monitor | speaker | table | watercraft | all |
|---|---|---|---|---|---|---|---|---|---|---|---|---|---|---|
| 3D-R2N2 [5] | 51.3 | 42.1 | 71.6 | 79.8 | 66.1 | 46.6 | 62.8 | 54.4 | 38.1 | 46.8 | 66.2 | 51.3 | 51.3 | 56.0 |
| OGN [27] | 58.7 | 48.1 | 72.9 | 81.6 | 70.2 | 48.3 | 64.6 | 59.3 | 39.8 | 50.2 | 63.7 | 53.6 | **63.2** | 59.6 |
| PSGN [8] | 60.1 | 55.0 | 77.1 | 83.1 | 74.9 | 54.4 | **70.8** | 60.4 | **46.2** | 55.2 | 73.7 | **60.6** | 61.1 | 64.0 |
| Ours (voxel tube network) | **67.1** | **63.7** | 76.7 | 82.1 | 74.2 | **55.0** | 69.0 | **62.6** | 43.6 | 53.4 | 68.1 | 57.3 | 59.9 | **64.1** |
| Ours (Matryoshka network) | 64.7 | 57.7 | **77.6** | **85.0** | **75.6** | 54.7 | 68.1 | 61.6 | 40.8 | 53.2 | 70.1 | 57.3 | 59.1 | 63.5 |

Table 2. **Single image 3D shape reconstruction on ShapeNet-core at $32^3$ resolution.** We report the mean IoU (%) per category, and the average IoU over all categories. Our networks outperform all voxel decoder baselines and are competitive with the more complex PSGN.

right). The encoder starts with 2 convolution layers with interleaved batch normalization and ReLU nonlinearity to produce 8 initial feature channels while keeping the input resolution. It is further composed of residual modules [12] that down-sample the input image to a resolution of $4 \times 4$ while linearly increasing the number of feature channels to 512 (257 for experiments on shapes of a single category and small resolution, *i.e.* our ablation study). Each residual module contains two $3 \times 3$ convolutions, with batch normalization and a ReLU nonlinearity before each convolution. Downsampling while increasing the number of feature channels is done in the first convolution layer of each residual module. Modules altering the spatial resolution alternate with modules operating at the same resolution.

The decoder upsamples again until the desired output resolution is reached. Mirroring the encoder, the decoder is also composed of residual modules and decreases the number of feature channels linearly. For upsampling, the first convolution is replaced with a transposed convolution. In all our experiments, we trained using Adam [16] using an initial learning rate of 0.001 and $\beta_1 = 0.9, \beta_2 = 0.999$, and varied the schedule for dropping the learning rate with the dataset size. We refer to the supplemental for more details.

## 4. Evaluation

To assess the performance of our geometry prediction approaches at different tasks, compare them to prior work, and study the influence of loss functions and network architectures, we a use a common subset [5, 8, 27] of ShapeNet-core [4]. The subset consists of nearly 50000 3D shapes divided into 13 major categories. For all experiments, we report the intersection over union (IoU) in %.

### 4.1. Reconstruction from a single view

**Comparison to prior work.** For evaluating the performance of our networks in reconstructing 3D shape from a single RGB image, we compare to 3 recent approaches on the ShapeNet-core dataset using the renderings, dataset split, and ground truth voxel representations provided by Choy *et al.* [5]. The renderings feature images of size $137 \times 137$ and a uniform sampling of viewpoints. The voxel representations are of size $32 \times 32 \times 32$. As preprocessing step, we crop input images to $128 \times 128$ and shuffle color channels randomly during training. We train a single network for all shape categories. We compare to different representative approaches for predicting 3D shapes: *(1)* 3D-R2N2 [5] features a dense 3D voxel decoder and an LSTM to enable reconstruction from one or multiple views; *(2)* Octree Generating Networks (OGN) [27] operate on octrees to exploit sparsity of occupancy maps; *(3)* Point Set Generation Networks (PSGN) [8] predict a point cloud using a stacked hourglass network, a volume prediction network, and a voxel-based post-processing network.

We show results in Tab. 2. Although conceptually simpler, the dense *voxel tube image* version of our network outperforms all voxel decoder-based approaches and is on par with PSGN, which uses a more complex multi-stage (multi-network) architecture. Moreover, it is not clear if PSGNs scale to higher resolutions, whereas this is easily possibly for our networks (see below). Interestingly, the sparse Matryoshka version of our network, which predicts *nested shape layers*, performs only slightly worse than its dense counterpart and clearly outperforms all voxel decoder baselines. This demonstrates the power of our compact image-based representation for 3D shape.

| Method | Category | $32^3$ | $64^3$ | $128^3$ | $256^3$ |
|---|---|---|---|---|---|
| OGN [27] | car | 64.1 | 77.1 | 78.2 | 76.6 |
| Ours (Matryoshka) | car | **68.3** | **78.4** | **79.4** | **79.6** |
| | airplane | 36.7 | 48.8 | 58.0 | 59.6 |
| | table | 38.6 | 42.3 | 43.5 | 41.3 |

Table 3. **Single image 3D shape reconstruction for high resolutions.** We report IoU (in %) between predictions at several resolutions and ground truth shapes at $256^3$. Predictions at lower resolution are up-sampled to $256^3$.

**Reconstructing higher resolutions.** Low-resolution occupancy maps are naturally limited to a low level of detail they can represent. To assess the performance of our Matryoshka network at reconstructing shapes at high resolution, we compared it to Octree Generating Networks [27], which are representative for Octree-based approaches. We follow the experimental setup of Tatarchenko *et al.* [27] and predict 3D shapes from ShapeNet-cars at resolutions of $32^3$, $64^3$, $128^3$, and $256^3$ given a single RGB input image. We then up-sample the predictions to a resolution of $256^3$ voxels and compute the intersection over union with the ground truth shapes at that resolution. For fair comparison, we use dataset split and ground truth shapes provided by Tatarchenko *et al.* [27]. Furthermore, we provide results for 2 additional classes from ShapeNet-core, which pose different challenges; while the airplane class consists of shapes with intricate structure, the table class contains the most samples. We report quantitative results in Tab. 3 and show qualitative examples in Fig. 5. We find that both methods predict more accurate shapes at higher resolutions. However, OGN's performance saturates at $128^3$ due to the high complexity of the car class with 7496 samples. For our method we only observe this effect in the even more complex table category (8509 samples). For all resolutions, the proposed method clearly outperforms the octree-based approach despite being based on standard 2D networks, which can be easily implemented in all popular frameworks. The benefits of higher resolutions are observed best for the airplane class, which shows the biggest relative improvements.

### 4.2. Ablation studies

To better understand the contribution of individual components to the overall performance of our networks, we examine different base architectures and loss functions. For our ablation study we use the dataset split and renderings ($64 \times 64$ pixels) from Yan et al. [33], taken from the same 24 viewpoints for each object. For the study of loss functions we train one network per class and for the study of network architectures, we train one network for all classes.

**Network architectures.** We investigate several network architectures that are known to perform well for dense 2D pre-diction tasks. As memory consumption is a dominating factor in the choice of a suitable architecture, we modified all architectures to fit within 3 GB of GPU memory when predicting shapes at a resolution of $32^3$ with a mini-batch size of 128. This leaves sufficient memory budget for scaling up any architecture to higher output resolutions on a single GPU. Since some of the base architectures (ResNet [12], DenseNet [14]) are designed to operate on images, but to produce a single class label, they require adaptation to generate dense output. In the interest of space, we defer architectural details to the supplementary material. We take our voxel tube network as ResNet-inspired baseline. Removing residual connections yields an Encoder/Decoder or Deconvolution Network [19]; the introduction of connections between layers of the same spatial resolution to skip varying sequences of down- and up-sampling forms a U-Net inspired network, *c.f.* [23]. We report results in Tab. 4. Across all categories, the ResNet-inspired architecture outperforms all other networks with a significant margin. Note that, *e.g.*, octree-based decoders, in contrast to our approach, cannot take advantage of this as easily.

| Base architecture | car | chair | table | mean |
|---|---|---|---|---|
| Encoder/decoder [19] | 73.0 | 52.5 | 57.0 | 60.8 |
| U-Net [23] | 74.2 | 54.8 | 58.8 | 62.6 |
| ResNet [12] | **75.6** | **56.8** | **59.1** | **63.8** |
| DenseNet [14] | 72.3 | 49.4 | 55.8 | 59.2 |

Table 4. **Evaluation of base architectures.** Across all categories, the ResNet-inspired architecture outperforms all other networks with a significant margin.

| Loss function | car | chair | table | mean |
|---|---|---|---|---|
| Binary cross-entropy | 75.9 | 57.8 | 58.2 | 64.0 |
| $\mathcal{L}_1$ | 73.6 | 57.2 | 57.4 | 62.7 |
| $\mathcal{L}_2$ | 76.4 | 58.0 | 58.7 | 64.4 |
| Cosine similarity $\mathcal{L}_C$ | 75.7 | **58.4** | 59.3 | 64.5 |
| Approx. IoU $\mathcal{L}_{\text{IoU}}$ | **76.3** | 58.3 | **59.5** | **64.7** |

Table 5. **Influence of loss functions.** We report the IoU for our voxel tube network trained with several loss functions on the car, chair, and table categories.

**Loss functions.** To assess how specific loss functions affect the reconstruction quality for our voxel tube networks, we evaluate the binary cross entropy, $\mathcal{L}_1$-norm, $\mathcal{L}_2$-norm, the negative cosine similarity (Eq. 7), and the negative intersection over union (Eq. 8). We report results in Tab. 5. We find that the binary cross entropy is a strong baseline, but performs worse than all other evaluated loss functions except for the $\mathcal{L}_1$-norm, which consistently performs worst. Since the evaluated architecture constrains activations in the final layer between 0 and 1, a robust loss is less important. For all categories, the two proposed losses perform best.

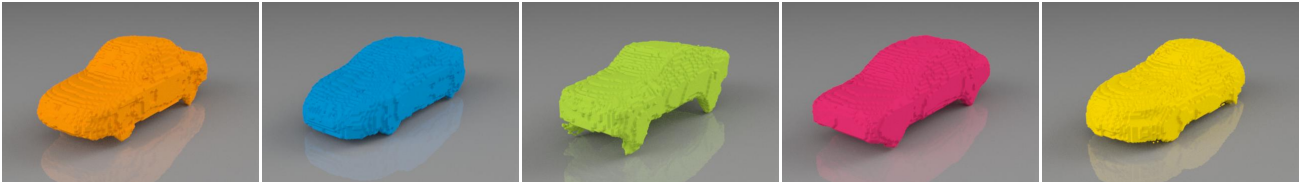

Figure 6. **Sampling shapes.** By supplying the SfSS-decoder with Gaussian noise, we can draw varied samples from the car distribution.

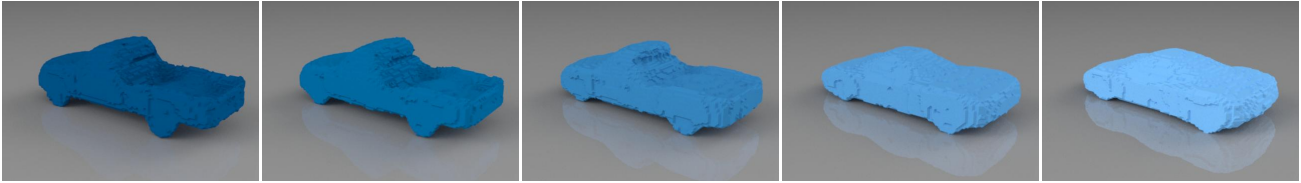

Figure 7. **Shape interpolation.** Linearly interpolating the descriptors we feed to the SfSS-decoder produces plausible interpolations of the generated shapes.

### 4.3. Other applications

**Shape from ID.** To assess the ability of our method to represent highly complex datasets, we follow Tatarchenko *et al.* [27] and predict shapes from their Blendswap dataset at $512^3$ voxels from a high-level shape ID. We find that our method is able to reconstruct the dataset at a similar quality level (97.8% IoU) as OGN [27] (96.9% IoU), but in contrast to [27] using a 2D representation alone.

**Shape from shape similarity and shape generation.** We aim to assess our model's ability to reconstruct shapes from high-level information without relying on a specific image encoder architecture. To that end, we train our network to generate shapes from a high-dimensional descriptor that captures shape similarities within a semantic category. We construct the descriptor by computing a pairwise similarity matrix of 3D models such that an entry at $(i, j)$ represents the intersection over union between models $i$ and $j$ of resolution $32^3$ in the ShapeNet-cars dataset. Reducing the dimensionality of the matrix with PCA while retaining 95% of the variance and removing duplicates yields 2424-dimensional descriptors for 7426 remaining shapes. Trained on 80% of the descriptors to generate shapes at $128^3$ voxels resolution, our Matryoshka network reaches a mean intersection over union of 81.1% on the held out shapes. This Shape-from-Shape-Similarity (SfSS) decoder can also be used for interpolating between shapes (Fig. 7) and to synthesize new shapes by supplying a random noise vector. As shown in Fig. 6, samples drawn from the model are quite diverse (*c.f.* Fig. 5 of [25] in contrast).

**Reconstruction from real images.** To show the applicability of our method to real-world images, we give a qualitative example in Fig. 8, *c.f.* supplemental for more examples.

**Shape from silhouette.** In the supplemental material, we additionally study the ability of our Matryoshka networks to reconstruct a 3D shape from a single silhouette image.

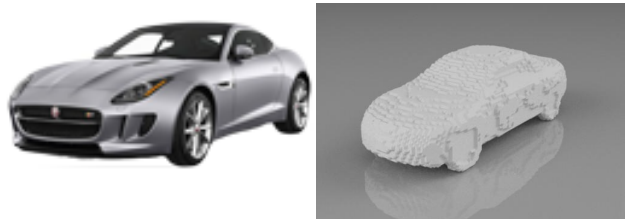

Figure 8. **Qualitative result for 3D shape reconstruction from real-world images.**

### 5. Conclusion

In this paper, we posed 3D shape reconstruction as a 2D prediction problem, allowing us to leverage well-proven architectures for 2D pixel-prediction. Both proposed networks clearly outperform dense voxel-based approaches at low resolutions. Our novel efficient encoding based on nested shape layers, furthermore, allows to scale our Matryoshka networks to handle reconstruction of shapes at a high resolution, while outperforming octree-based decoder architectures with a considerable margin, despite being based only on standard network layers. Applications to shape from ID and shape similarity, as well as shape sampling demonstrated the broad applicability of our approach.

The proposed shape layer encoding requires fewer than 5 layers even for high-resolution shapes. We consequently fix the maximum number of components in our experiments. To encode arbitrarily complex objects without requiring retraining, the required number of components could be predicted per individual object in a recursive formulation. We leave this and learning the shape fusion [21] for future work.

**Acknowledgments.** The research leading to these results has received funding from the European Research Council under the European Union's Seventh Framework Programme (FP/2007–2013)/ERC Grant agreement No. 307942.

# Appendix

## A. Network Architecture

Our networks consist of multiple residual modules (*c.f.* [12]) as shown in Fig. 9. Across all modules, we keep the kernel size $k$ and the stride $s$ constant as depicted. For all convolutions with a kernel size $k > 1$, we use reflective padding of size 1. Altering the spatial resolution and number of feature channels requires special handling of the identity pathway of the residual modules. For down-sampling (Fig. 9, middle), we simply drop every other pixel and initialize the added feature channels as zero using zero-padding. For up-sampling (Fig. 9, right), we use nearest neighbor interpolation and a $1 \times 1$ convolution to project the feature dimension. We experimented with more sophisticated up- and down-sampling alternatives, but found no significant benefits.

In all experiments with images as inputs, processing in our networks begins with a feature generation module, which produces an initial representation with $f_{in}$ feature channels. This module is equivalent to the residual module operating at constant resolution (Fig. 9, left), but with the first rectified linear unit and identity pathway removed. Each module is only parametrized by the number of feature channels added during down-sampling $\Delta f_\downarrow$ or removed during up-sampling $\Delta f_\uparrow$, and we pair each up-sampling and down-sampling module with a subsequent module of same resolution to form one residual block. Thus, we specify network architectures by a desired number of initial features $\hat{f}_{in}$, output features $\hat{f}_{out}$, features at the bottleneck $f_{inner}$, number of desired down-sampling blocks $d$ in the decoder, and residual blocks at the bottleneck $b$. We match the number of down-sampling blocks in the encoder with the number of up-sampling blocks in the decoder. We set it to 3 for an output resolution $s_{out} = 32$ and increase it by 1 for every doubling of $s_{out}$. If input and output resolutions are different, we add $d_i = \log_2 s_{in} - \log_2 s_{out}$ down-sampling blocks or $d_o = \log_2 s_{out} - \log_2 s_{in}$ up-sampling blocks accordingly. For all networks, we scale input images to powers of 2. We compute the number of feature channels to add for each down-sampling block as

$$\Delta f_\downarrow = \left\lfloor \frac{f_{inner} - \hat{f}_{in}}{d_i + d} \right\rfloor, \quad (10)$$

and adjust the number of initially generated features as

$$f_{in} = \hat{f}_{in} + (f_{inner} - \hat{f}_{in}) \bmod \Delta f_\downarrow \quad (11)$$

to obtain integral numbers for the number of feature channels. Analogously, we compute the number of feature channels added per up-sampling block as

$$\Delta f_\uparrow = \left\lfloor \frac{f_{inner} - \hat{f}_{out}}{d_o + d} \right\rfloor. \quad (12)$$

To obtain predictions with the desired number of output channels (equaling $s_{out}$ for voxel tube networks and the number of shape layers $\times 6$ for Matryoshka networks) we simply add a 2D convolution with kernel size 1 as final layer to our networks. We summarize the architectures and training schedules used in the individual experiments in Tab. 7.

For a batch size of 128, we start with a learning rate of 0.001 and reduce it by a factor of 10 after *drop* epochs. For any different batch size, we scale the learning rate accordingly. All models were trained on a single GPU.

For the ablation studies, we used a voxel tube network as summarized in the penultimate row of Tab. 7. Since the networks for the shape-from-silhouette task and the ablation studies were trained on renderings of smaller resolution and on a smaller number of categories, we roughly halved the number of feature channels at the bottleneck (setting it to 257 for an integer $\Delta f_\uparrow$).

For the study on network architectures, we refer to the voxel tube network as described above as ResNet-based network. We remove the identity pathways from all residual modules to obtain an Encoder/decoder network, and add skip connections between layers of same spatial resolution to obtain a U-Net, *c.f.* [23]. To adapt the number of feature channels for the skip connections, we use a $1 \times 1$ 2D convolution akin to the identity path of the up-sampling module in Fig. 9. The DenseNet-inspired version (*c.f.* [14]) of our voxel-tube network consists of 7 dense blocks (B), 2 up-transitions (U) and 3 down-transitions (D) arranged as BDBDBDBBUBUB. Each dense block contains 2 dense layers with an expansion factor of 16. For the down-transitions we halve the spatial resolution while keeping the number of feature channels constant. For the up-transitions we double the spatial resolution and halve the number of feature channels.

## B. More Results

**Shape from silhouette.** We investigate the performance of our Matryoshka network on the task of reconstructing a 3D shape from a single silhouette image. To that end, we reconstruct the shapes of the 3 categories with the most examples (chair, car, table) from ShapeNet-core with the dataset split and shapes from Choy *et al.* [5]. We obtain silhouettes from the alpha-channels of the renderings of Choy *et al.* As can be seen in Tab. 8, the network performs much better on cars than on tables or chairs. This can be attributed to the approximately convex shape of cars, which makes their silhouette a very effective cue for the overall shape. Compared to the easier setting of reconstructing shapes from a color image, the network performs remarkably well. Note, however, that the network for predicting shapes from color images was trained in a category-agnostic way, making the prediction considerably harder.

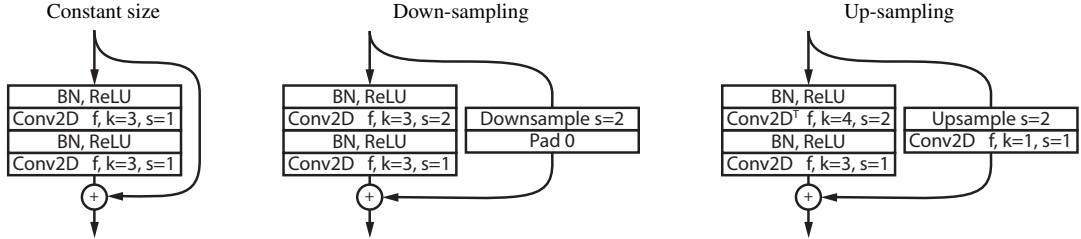

Figure 9: **Residual modules.** Each residual module consists of Batch normalization (BN) and Rectified Linear Unit (ReLU) layers followed by a 2D convolution, except for the up-sampling module where we replace the first convolution with a transposed convolution. The number of feature channels is denoted as $f$. Moreover, $k$ denotes the filter size and $s$ the stride.

| Network | $s_{in}$ | $s_{out}$ | batch size | epochs | drop | $\hat{f}_{in}$ | $d$ | $f_{inner}$ | $b$ | $\hat{f}_{out}$ |
|---|---|---|---|---|---|---|---|---|---|---|
| *ShapeNet-all* | | | | | | | | | | |
| Voxel tube | 128 | 32 | 128 | 45 | 15 | 8 | 3 | 512 | 1 | 32 |
| Matryoshka | 128 | 32 | 128 | 40 | 20 | 8 | 3 | 512 | 2 | 128 |
| *High resolution* | | | | | | | | | | |
| Matryoshka | 128 | 32 | 128 | 30 | 20 | 8 | 3 | 512 | 0 | 128 |
| Matryoshka | 128 | 64 | 128 | 30 | 20 | 8 | 4 | 512 | 3 | 128 |
| Matryoshka | 128 | 128 | 32 | 30 | 20 | 8 | 5 | 512 | 1 | 128 |
| Matryoshka | 128 | 256 | 8 | 30 | 20 | 8 | 6 | 512 | 0 | 128 |
| *Shape from Silhouette* | | | | | | | | | | |
| Matryoshka | 64 | 32 | 128 | 40 | 15 | 8 | 3 | 257 | 2 | 32 |
| *Shape from ID* | | | | | | | | | | |
| Matryoshka | 2 | 512 | 4 | 28K | 12K | – | 8 | 1 | 0 | 196 |
| *Ablation studies* | | | | | | | | | | |
| Voxel tube | 64 | 32 | 128 | 40 | 15 | 8 | 3 | 257 | 2 | 32 |
| *Shape from Similarity* | | | | | | | | | | |
| Matryoshka | 1 | 128 | 8 | 60 | 25 | – | 7 | 2424 | 0 | 128 |

Table 7: **Network architectures for individual experiments.** See text for a description of the network parameters.

| Category | car | chair | table | mean |
|---|---|---|---|---|
| Shape from silhouette | 86.7 | 53.2 | 58.8 | 66.2 |

Table 8: **Shape from silhouette** on ShapeNet-core.

**Real-world images.** To assess the performance of our proposed network for real world examples, we tested it on images from the Stanford Products Dataset [37] (chairs) and the web (cars). Qualitative examples are shown in Fig. 10 (chairs) and Fig. 11 (cars). In both cases, we trained a category-specific Matryoshka network to predict 3D shapes at $128^3$ resolution from a single image. For predicting chairs, we took the renderings of Choy *et al.* [5] and created ground truth shapes of higher resolution from the corresponding ShapeNet [4] models using *binvox* [36]. Since most images of cars found on the web are recorded from different camera positions than the renderings of Choy *et al.*, we re-rendered the car shapes from ShapeNet with random camera positions (focal length $\in [40mm, 90mm]$, azimuth $\in [0°, 360°]$, elevation $\in [0°, 25°]$) and environment maps collected from the web[1,2]. We find that Matryoshka networks generalize well to real-world imagery even when only trained on synthetic images. They are able to reconstruct thin structures (*e.g.*, the legs of the right-most chair in Fig. 10) and a wide variety of shapes (both Figs. 10 and 11).

**Synthetic images.** We show more results for predicting 3D shapes of high resolution in Figs. 13 (airplanes), 14 (chairs), and 12 (cars). The input images are renderings from Choy *et al.* [5] and the shapes have been converted to binary voxel grids using *binvox* [36]. The ground truth car shapes have been provided by Tatarchenko *et al.* [27]. Supporting the quantitative results from the main paper, learning to reconstruct 3D shapes at higher resolution produces much more accurate predictions, as can be seen for different resolutions in Fig. 12. Even for highly varied classes such as airplanes or chairs, Matryoshka networks produce high-quality reconstructions. Finally, we show qualitative examples of reconstructed shapes at low resolution from a voxel tube network and a Matryoshka network in Fig. 15. Both networks were trained on 13 categories from ShapeNet-core. Quantitative results for this experiment can be found in Tab. 2 in the main paper.

---
[1] http://www.hdrlabs.com/sibl/archive.html
[2] https://hdrihaven.com/

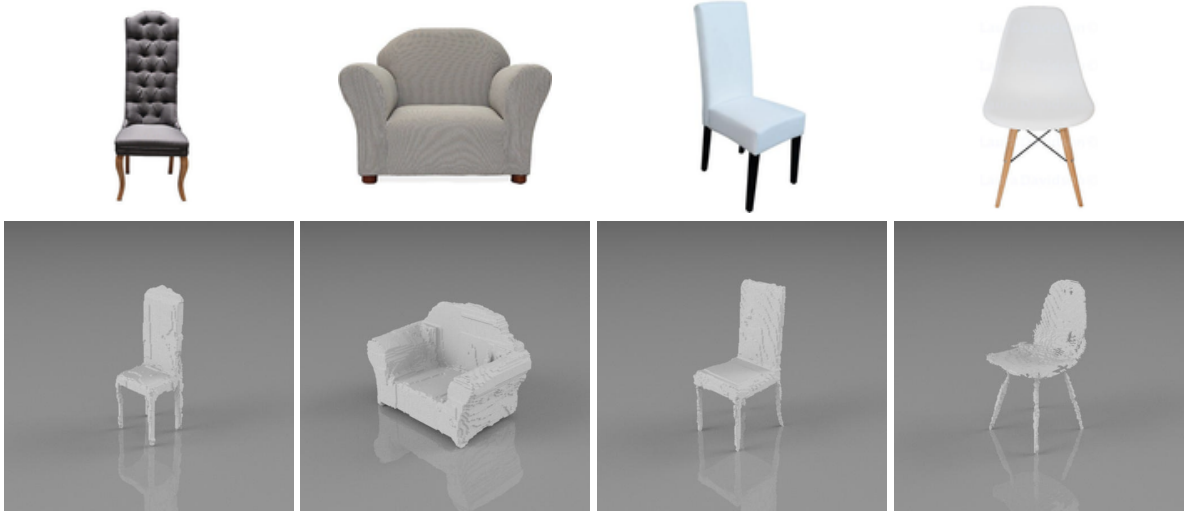

Figure 10: **Qualitative results at high resolution** ($128^3$) **for real-world images of chairs.** For a given input image (top row), our Matryoshka network predicts a 3D shape (bottom row).

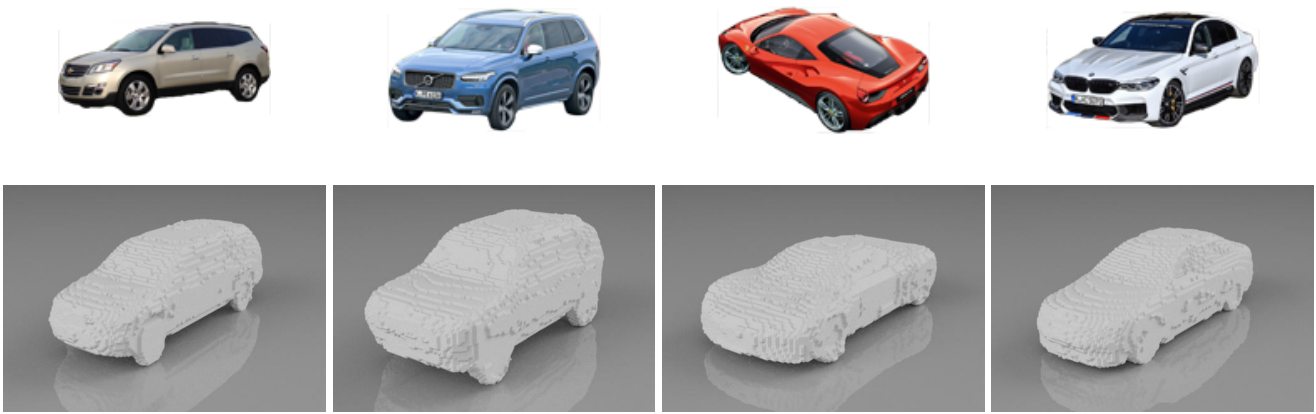

Figure 11: **Qualitative results at high resolution** ($128^3$) **for real-world images of cars.** For a given input image (top row), our Matryoshka network predicts a 3D shape (bottom row).

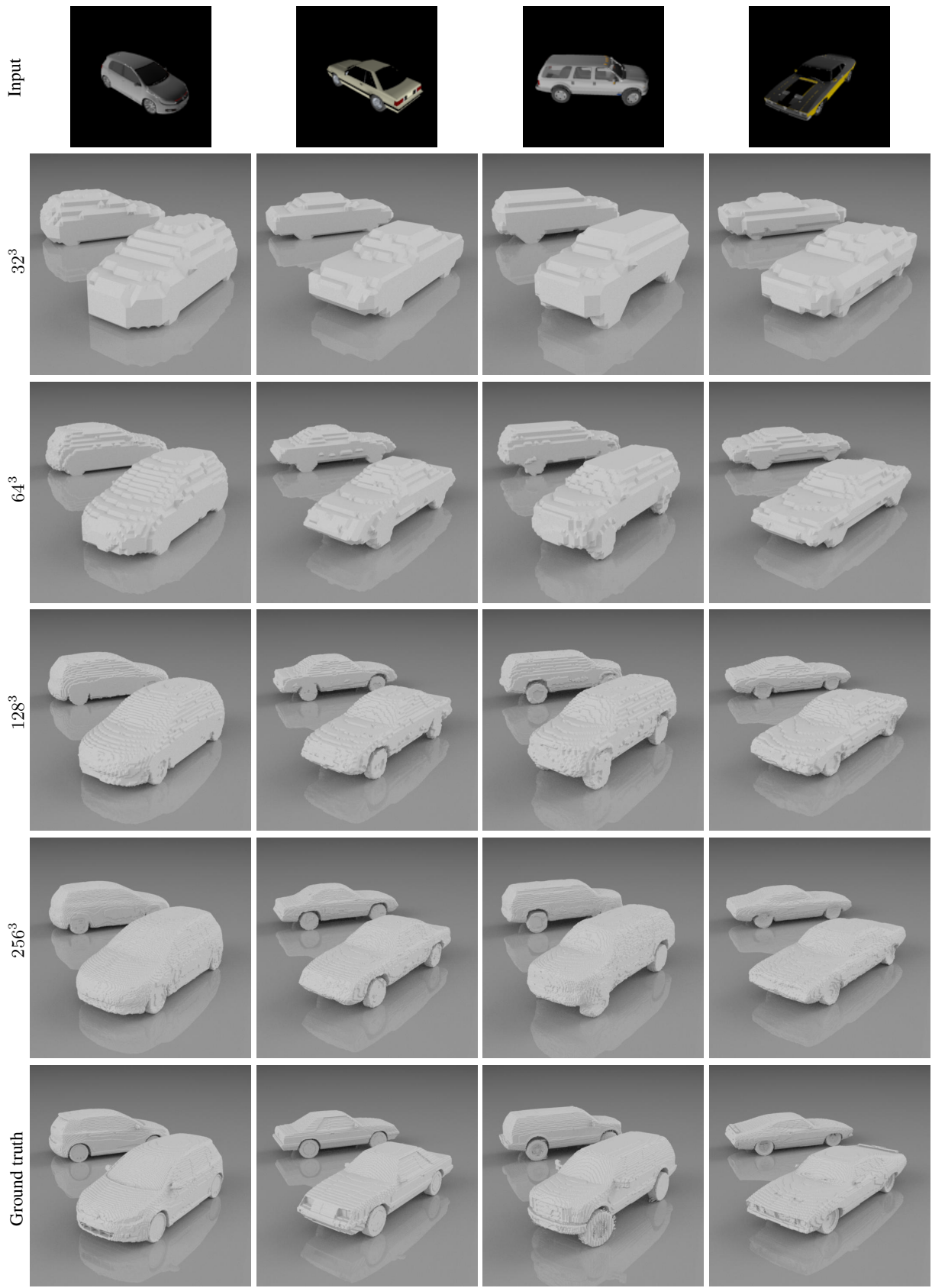

Figure 12: **Qualitative results at varying resolution.** We train Matryoshka networks to reconstruct 3D shapes from a single image rendered from ShapeNet models (top row) for output resolutions $32^3$, $64^3$, $128^3$, and $256^3$. The last row shows the ground truth shapes at $256^3$.

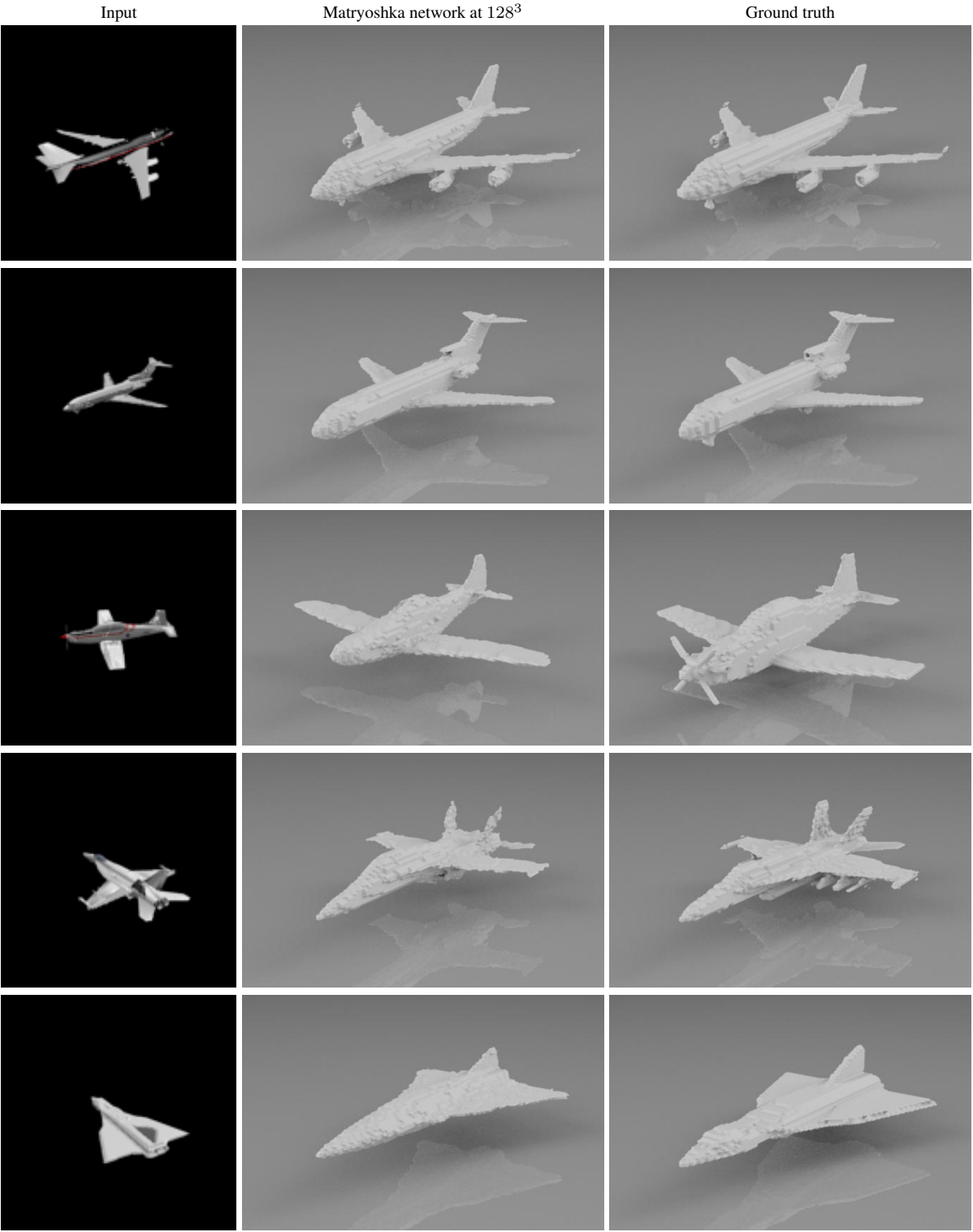

Figure 13: **Qualitative results at high resolution** ($128^3$) **for airplane images rendered from ShapeNet models.**

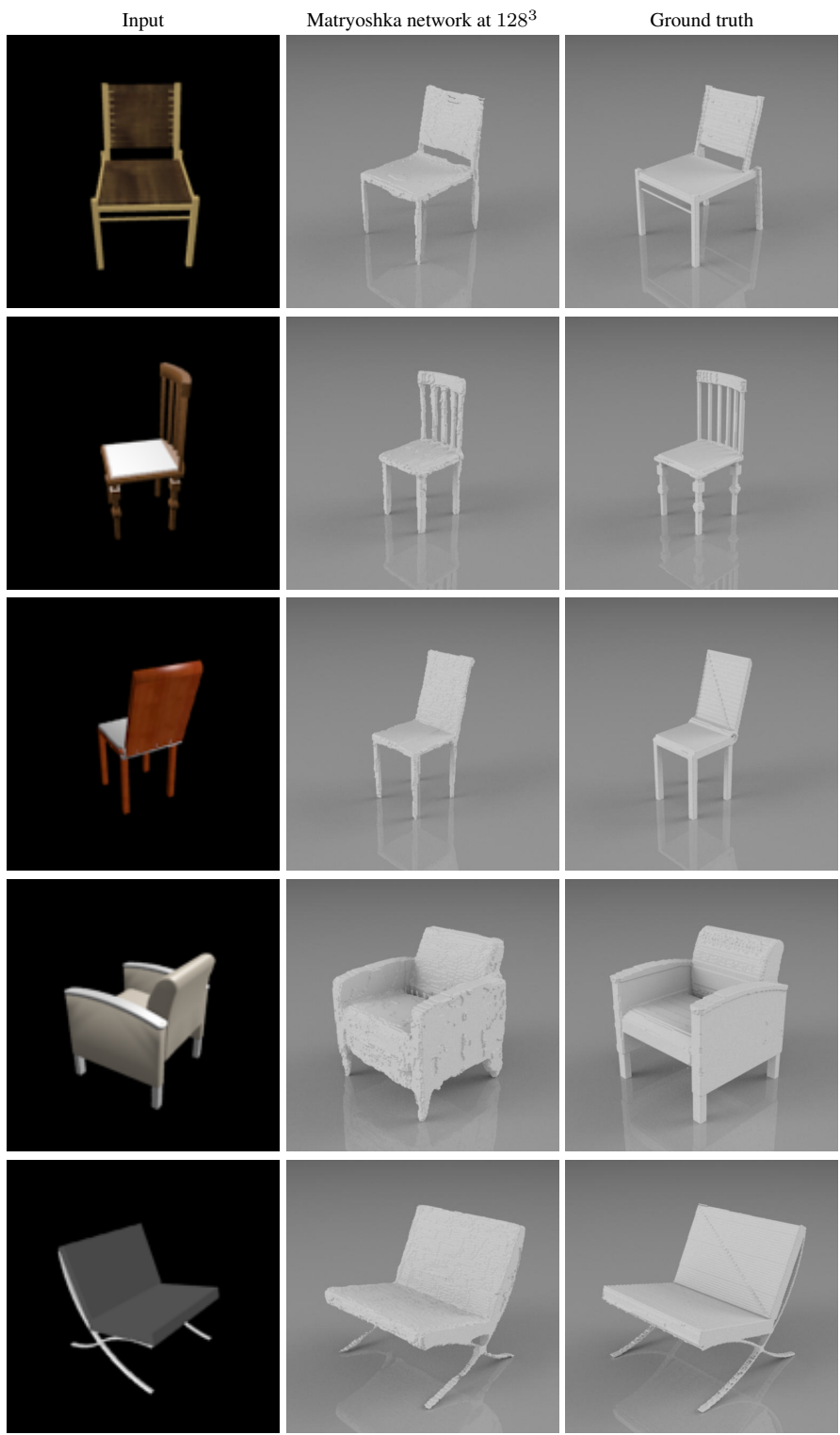

Figure 14: **Qualitative results at high resolution ($128^3$) for chair images rendered from ShapeNet models.**

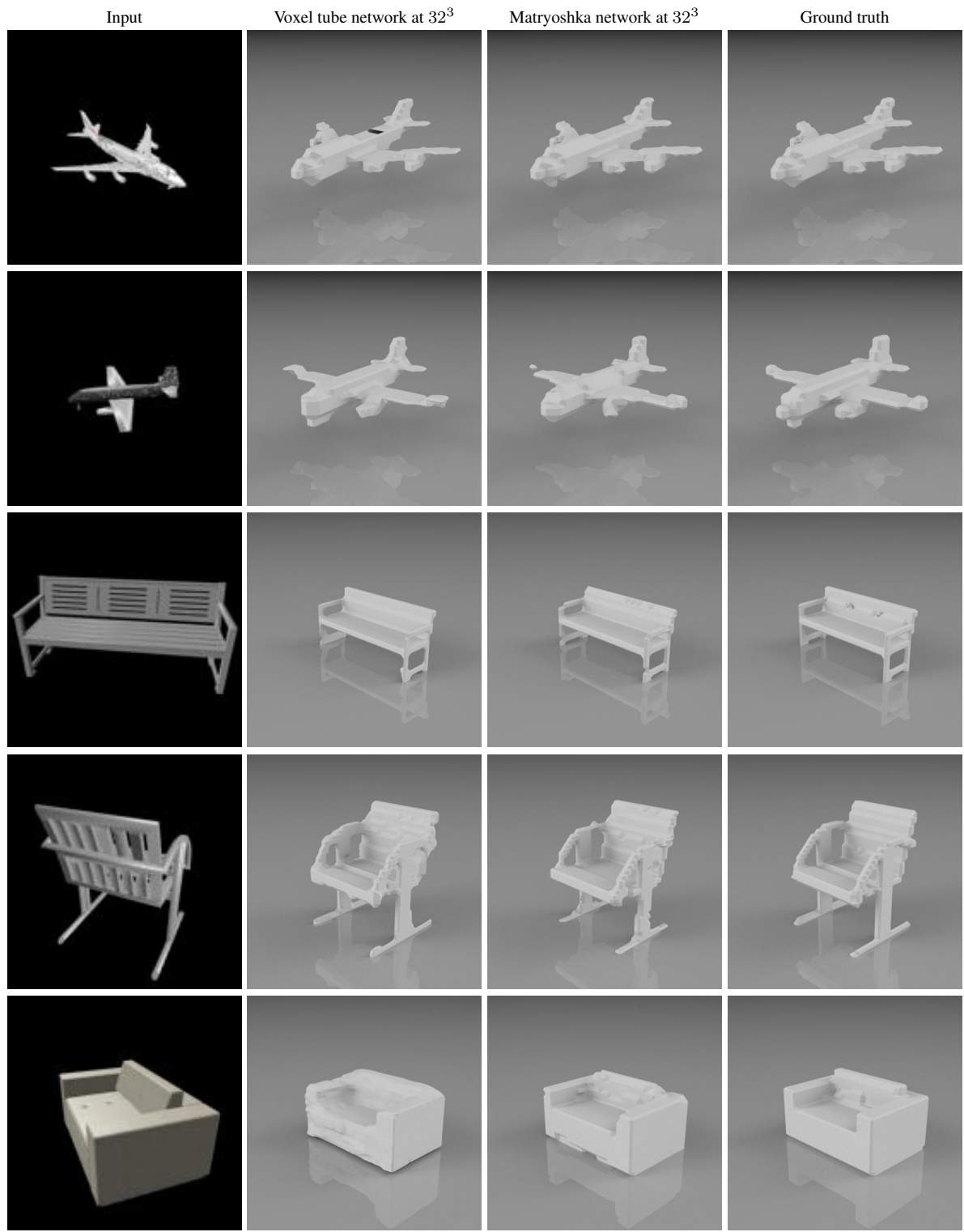

Figure 15: **Qualitative results at low resolution ($32^3$) for images rendered from ShapeNet models.** For input images (left-most row), we predict 3D shapes using a voxel tube network (2$^{nd}$ column) and a Matryoshka network (3$^{rd}$ column). Ground truth shapes are shown in the right-most column.

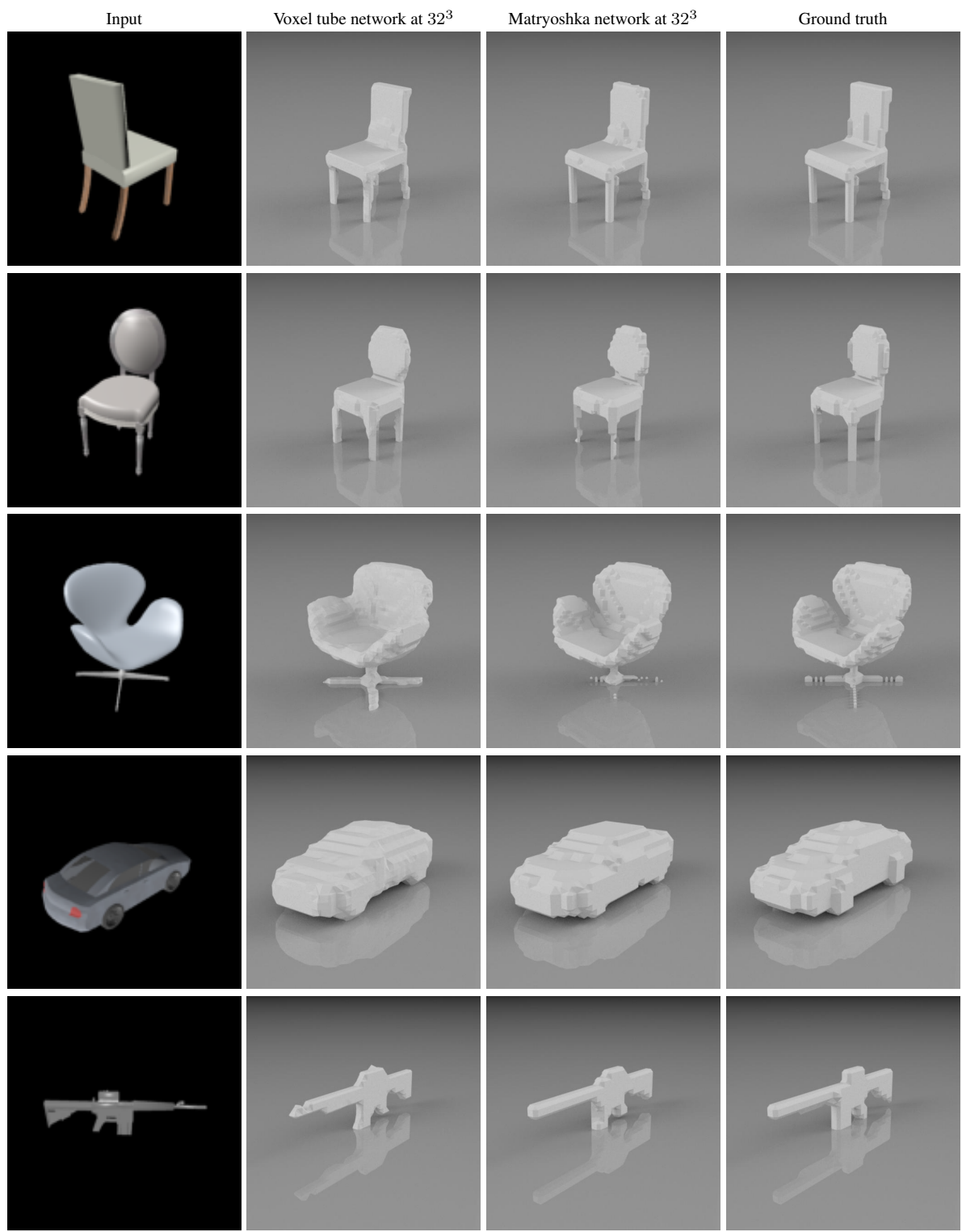

Figure 15: **Qualitative results at low resolution ($32^3$) for images rendered from ShapeNet models (continued)**.

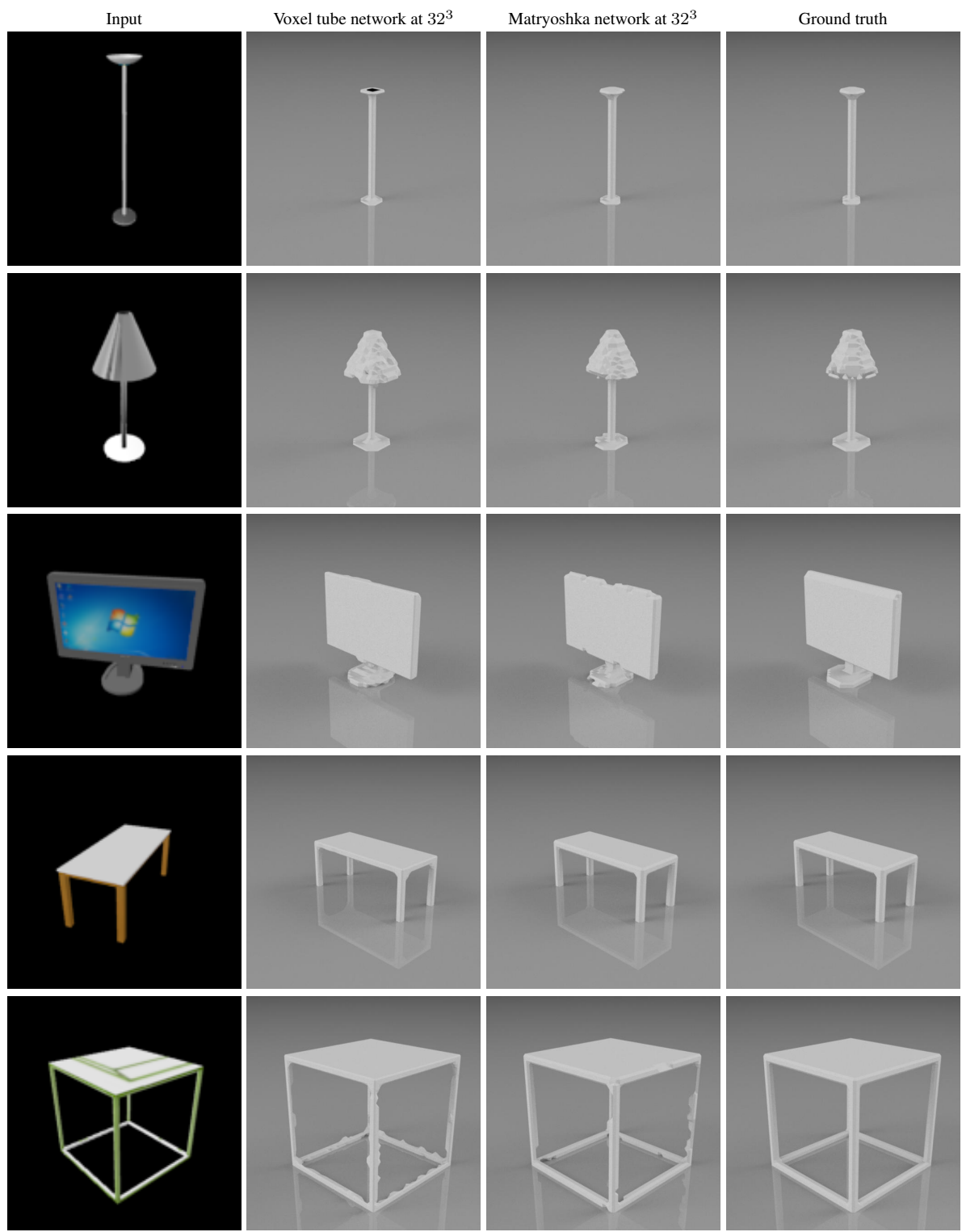

Figure 15: **Qualitative results at low resolution ($32^3$) for images rendered from ShapeNet models (continued)**.